\documentclass[conference]{IEEEtran}
\IEEEoverridecommandlockouts
\usepackage{cite}
\usepackage{amsmath,amssymb,amsfonts}
\usepackage{algorithmic}
\usepackage{graphicx}
\usepackage{textcomp}
\usepackage{booktabs}
\usepackage{wrapfig}
\usepackage{xcolor}
\usepackage{multirow}
\def\BibTeX{{\rm B\kern-.05em{\sc i\kern-.025em b}\kern-.08em
    T\kern-.1667em\lower.7ex\hbox{E}\kern-.125emX}}
\begin{document}

\title{
Leveraging CORAL-Correlation Consistency Network for Semi-Supervised Left Atrium MRI Segmentation
\\
}

\author{
\IEEEauthorblockN{Xinze Li\textsuperscript{1}, Runlin Huang\textsuperscript{1,2}, Zhenghao Wu\textsuperscript{1}, Bohan Yang\textsuperscript{1}, Wentao Fan\textsuperscript{1,3}, Chengzhang Zhu\textsuperscript{4}, Weifeng Su\textsuperscript{1,3}} 

\IEEEauthorblockA{
\textsuperscript{1} \textit{Department of Computer Science, BNU-HKBU United International College}
\textsuperscript{2} \textit{Hong Kong Baptist University} \\
\textsuperscript{3} \textit{Guangdong Provincial Key Laboratory of IRADS}
\textsuperscript{4} \textit{School of Humanities, Central South University} 
}
}

\maketitle

\begin{abstract}
Semi-supervised learning (SSL) has been widely used to learn from both a few labeled images and many unlabeled images to overcome the scarcity of labeled samples in medical image segmentation. Most current SSL-based segmentation methods use pixel values directly to identify similar features in labeled and unlabeled data. They usually fail to accurately capture the intricate attachment structures in the left atrium, such as the areas of inconsistent density or exhibit outward curvatures, adding to the complexity of the task. In this paper, we delve into this issue and introduce an effective solution, CORAL(Correlation-Aligned)-Correlation Consistency Network (CORN), to capture the global structure shape and local details of Left Atrium. Diverging from previous methods focused on each local pixel value, the CORAL-Correlation Consistency Module (CCM) in the CORN leverages second-order statistical information to capture global structural features by minimizing the distribution discrepancy between labeled and unlabeled samples in feature space. Yet, direct construction of features from unlabeled data frequently results in ``Sample Selection Bias'', leading to flawed supervision. We thus further propose the Dynamic Feature Pool (DFP) for the CCM, which utilizes a confidence-based filtering strategy to remove incorrectly selected features and regularize both teacher and student models by constraining the similarity matrix to be consistent. Extensive experiments on the Left Atrium dataset have shown that the proposed CORN outperforms previous state-of-the-art semi-supervised learning methods.
\end{abstract}


\section{Introduction}
Left atrium receives oxygenated blood from the pulmonary circulation that is then delivered to the left ventricle and then into the systemic circulation. Precise segmentation of the left atrium in cardiac MRI is crucial for diagnosing and monitoring heart conditions \cite{xiong2021global}. However, current challenges of direct segmentation of left atrium MRI images are due to attenuated contrast and its relatively small chamber and thin wall compared with other cardiac structures \cite{peng2016review}. Thereby, segmentation of left atrium is an integral part of clinical practice on heart disease.

Medical image segmentation has seen significant advancements with the advent of deep learning. However, the reliance on large volumes of annotated data presents a bottleneck due to the scarcity of labeled medical images and the extensive effort required for annotation. Semi-supervised learning approaches \cite{Wang2022,Luo_Chen_Song_Wang_2021,bioengineering10070869} emerge as a promising solution to this challenge by leveraging both labeled and unlabeled data.

Most current research focuses on cross-teaching paradigms \cite{semantic,Mutual,Xie_2020_CVPR}, utilizing cross-pseudo supervision methods. In these methods, two identically structured but independently initialized networks mutually supervise each other by generating pseudo labels. Based on cross-teaching paradigms, some studies employ strategies such as consistency regularization \cite{HLS,CCT} and entropy minimization \cite{Pseudo,Meta,Atrium} to ensure model robustness against data perturbations and to promote high-confidence predictions across classes, respectively. These semi-supervised methods have achieved promising progress.

Previous methods\cite{Cyclic,Luo_Chen_Song_Wang_2021,Shape} primarily relied on pixel-level supervision. This involves annotating each pixel with a label, which may only capture very local information and often neglect the global shape, contributing to the production of wrong pseudo-labels. Moreover, the accumulation of wrong pseudo-labels will mislead to the model. To address these issues, we propose the CORAL-Correlation Consistency Network (CORN), which leverages the second-order statistical features to extract both global structures and local details \cite{sun2017correlation,SOS2,SOS3}, dynamically aligning the distribution of labeled and unlabeled data to accurately capture left atrium structures.

\begin{figure*}[htbp] 
\centering 
\includegraphics[width=0.99\textwidth]{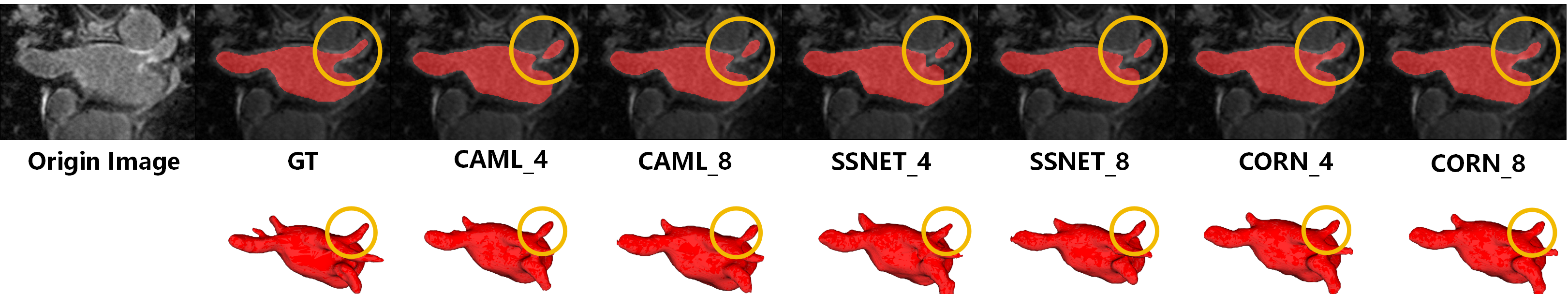} 
\caption{Comparative Visualization of Segmentation: Previous Methods vs. CORN. Numerical subscripts denote the results from the model trained with the corresponding number of labeled samples.}
\label{Fig.visual} 
\end{figure*}

While traditional methods frequently encounter ``segmentation breakage'', our CORN network demonstrates robustness. Particularly in edge regions, CORN surpasses conventional methods by delivering enhanced details and producing shapes that more accurately reflect the ground truth (GT). This superior performance is clearly illustrated in Fig. \ref{Fig.visual}.

The issue of inaccurate supervision in semi-supervised segmentation, primarily caused by the misrepresentation of features extracted from unlabeled data and inappropriate feature selection, often leads to ``Sample Selection Bias''. To mitigate this problem, we introduce the Dynamic Feature Pool (DFP). This module deploys a confidence-based filter to discard features associated with low segmentation confidence and incorrect predictions, and it uses a consistency strategy for model regularization.

This strategy predominantly retains consistent features from the dual encoder, which are deemed significant, thereby enhancing model performance and reliability. In terms of robustness, we consider that some inconsistent features may represent rare characteristics. Therefore, these features are also filtered in for comparison.

When dynamically updating features in the pool, we employ a fusion strategy to incorporate more traits learned by the dual model from the unlabeled data. This approach ensures a more comprehensive and accurate representation of data, further improving the model's performance.

Our contributions are summarized as follows: (1) We introduce the CORAL-Correlation Consistency Network (CORN), a novel approach that leverages second-order statistical features to capture the complex organ structure in medical images, with attention to both global structures and local details. (2) The Dynamic Feature Pool (DFP) employs a confidence-based filtering and consistency strategy, effectively mitigating the negative impact of inaccurate features derived from unlabeled data. (3) Extensive and detailed experiments demonstrate the effectiveness of our approach over the previous state-of-the-art on the Left Atrium dataset.

\section{Method}

\begin{figure}[htbp] 
\centering 
\includegraphics[width=0.99\columnwidth]{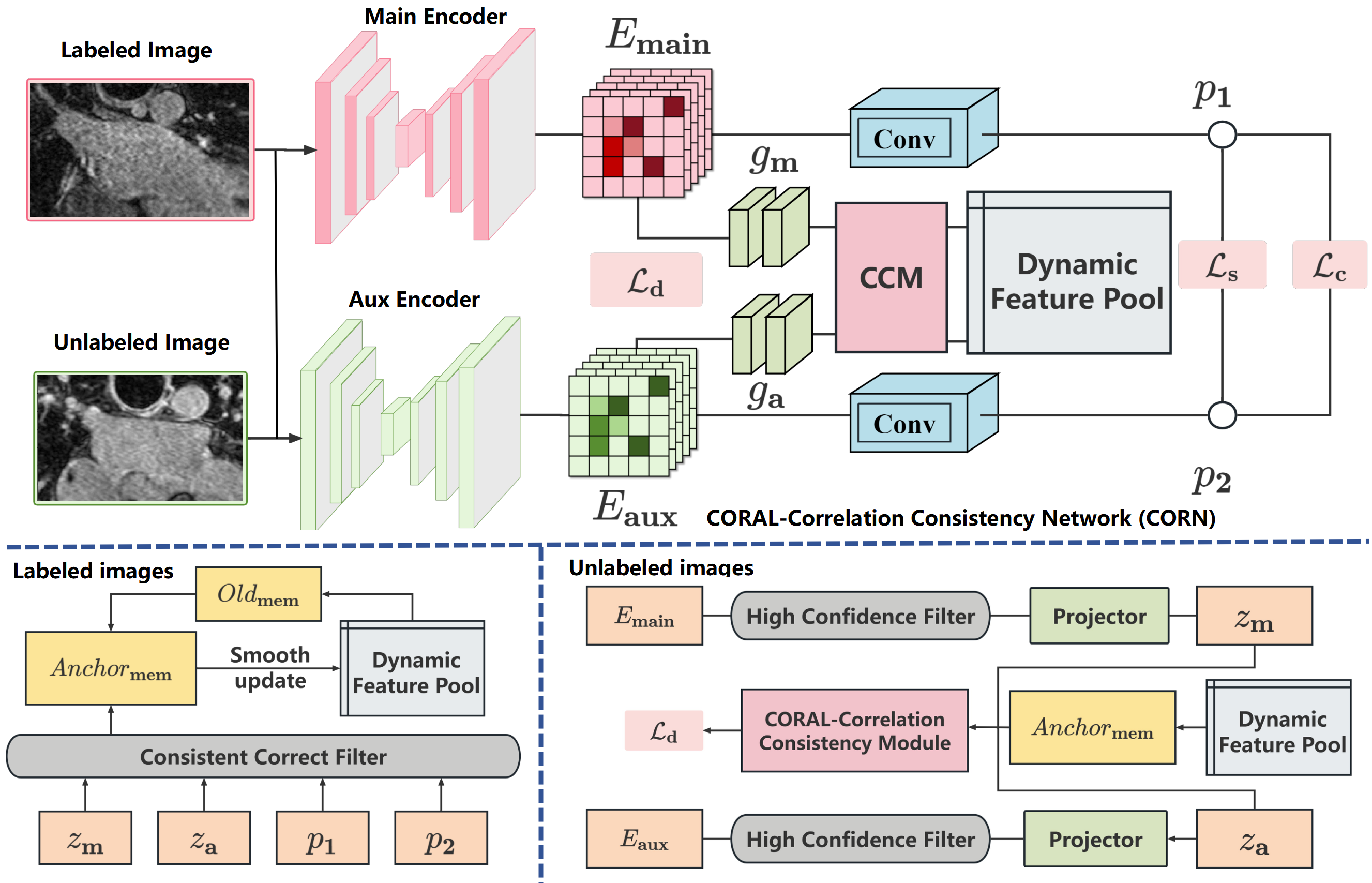} 

\caption{Overview of our segmentation network. The CORN unifies the Dynamic Feature Pool with a consistency filtering strategy to capture global and local features from both labeled and unlabeled images.}
\label{Fig.main_structure} 
\end{figure}

Fig. \ref{Fig.main_structure} illustrates the comprehensive architecture of our segmentation network: CORAL-Correlation Consistency Network
(CORN). Our network employs dual parallel encoders for feature extraction and predicting segmentation results. Additionally, receiving the projection head of segmentation results as the input, Dynamic Feature Pool (DFP) is employed to make feature filtering and dynamically update features in the pool with a fusion mechanism. Later, the output of DFP, which are selected as high-quality features for regulation of unlabeled data representation, is transferred to CORAL-Correlation Module (CCM), where statistical distributions of features between source (labeled data) and target (unlabeled data) domains are aligned. This network enables the semi-supervised learning process to explore the correlation within samples effectively and the features among labeled and unlabeled data, contributing to solving the segmentation of ambiguous boundaries. This network is well devised to identify the anchor points by exploring the synergy between the labeled and unlabeled data. Further insights into CORN will be introduced in subsequent sections. 

The total loss of the CORAL-Correlation Consistency Network can be formulated as:
    \begin{equation}
        L = \mathcal{L}_s + \lambda_cl_c + \lambda_dl_d 
    \end{equation}
where $\mathcal{L}_s$ and $l_c$ are the cross-supervised loss completed in the \textit{Cross Pseudo Supervision (CPS)} network \cite{semantic}, and $l_d$ represents the proposed CORAL-Correlation Consistency loss. $\lambda_c$ and $\lambda_d$ control the the weights of $l_c$ and $l_d$ respectively. While supervised loss $\mathcal{L}_s$ is applied exclusively to the labeled data, all samples contribute to the formation of cross-supervised learning.

The supervision loss $\mathcal{L}_s$ \cite{semantic}, is based on the standard pixel-wise cross-entropy loss on the labeled images over the two parallel segmentation networks: 

\begin{equation}
    \mathcal{L}_s=\frac{1}{\left|\mathcal{D}^l\right|} \sum_{\mathbf{X} \in \mathcal{D}^l} \frac{1}{S} \sum_{i=0}^{S}\left(\ell_{c e}\left(\mathbf{p}_{1 i}, \mathbf{y}_{1 i}^*\right)\right. \\
\left.+\ell_{c e}\left(\mathbf{p}_{2 i}, \mathbf{y}_{2 i}^*\right)\right)
\end{equation} $\mathcal{D}^l$ denotes the given of $N$ labeled images, with $S = W \times H$ as the image size. $\mathbf{y_{1 i}^*}$ and $\mathbf{y_{2 i}^*}$ indicate the ground truths, while $\mathbf{p_{1 i}}$ and $\mathbf{p_{2 i}}$ represent the segmentation confidence map vectors at each position $i$ from the two parallel segmentation networks after the softmax normalization. 

The cross pseudo supervision loss $l_c$ is bidirectional. Like the supervision loss, the pixel-wise one-hot label map $\mathbf{y}{1}$ from the main encoder $E{main}$ supervises the confidence map $\mathbf{p}{2}$ from the auxiliary encoder $E{aux}$, and $\mathbf{y}{2}$ from $E{aux}$ supervises $\mathbf{p}{1}$ from $E{main}$. The cross pseudo supervision loss on unlabeled data is defined as:

\begin{equation}
    l_c=\frac{1}{\left|\mathcal{D}^u\right|} \sum_{\mathbf{X} \in \mathcal{D}^u} \frac{1}{S} \sum_{i=0}^{S}\left(\ell_{c e}\left(\mathbf{p}_{1 i}, \mathbf{y}_{2 i}\right)\right. \\
\left.+\ell_{c e}\left(\mathbf{p}_{2 i}, \mathbf{y}_{1 i}\right)\right)
\end{equation} where $\mathcal{D}^u$ denotes the $M$ unlabeled images.

\subsection{CORAL-Correlation Consistency Module}
The CORAL-Correlation Consistency Module (CCM) introduces a novel regularization method leveraging second-order statistical information \cite{sun2017correlation}. The CCM computes a similarity matrix by comparing features of unlabeled data with anchor point features from labeled data, aimed to align the feature distributions between labeled and unlabeled data. To enhance the robustness of CORN, we maintain consistent correlation across another branch for diverse unlabeled features. This involves using a CORAL-like method to calculate the correlation matrix, formulating the similarity distribution between anchor point features and unlabeled features.

Let the projection heads $g_m$ and $g_a$ be connected to the main encoder $E_{main}$ and auxiliary encoder $E_{aux}$ independently, with $z_m \in \mathbb{R}^{m \times c'}$ and $z_a \in \mathbb{R}^{m \times c'}$ symbolizing two distinct sets of feature embedding vectors, derived from their respective projected outputs. Here, $m$ is the number of sampled features, and $c'$ is the dimension of these projections. It is essential to recognize that both $z_m$ and $z_a$ originate from matching locations within unlabeled samples. Assuming $z_p \in \mathbb{R}^{n \times c'}$ denotes the anchor point sampled from labeled data, where $n$ is the count of these sampled features,
We define the CORAL loss as the distance between the second-order statistics (covariance) of the source ($z_p$) and target ($z_m$) features:

\begin{equation}
    \mathbf{CORAL}(z_p, z_m) = \frac{1}{4{c'}^2} \| cov(z_p) - cov(z_m) \|_F^2
\end{equation}

Here, $cov(.)$ denotes the covariance matrix, and $\|.\|_F$ is the Frobenius norm. The following formulation demonstrates the computation process of the CORAL-Correlation matrix:

\begin{equation}
\resizebox{\columnwidth}{!}{$
    CORR_{mp} = \begin{bmatrix}
        \mathbf{CORAL}(z_{p_1}, z_{m_1}) & \cdots & \mathbf{CORAL}(z_{p_n}, z_{m_1}) \\
        \mathbf{CORAL}(z_{p_1}, z_{m_2}) & \ddots & \mathbf{CORAL}(z_{p_n}, z_{m_2}) \\
        \vdots & \vdots & \vdots \\
        \mathbf{CORAL}(z_{p_1}, z_{m_m}) & \cdots & \mathbf{CORAL}(z_{p_n}, z_{m_m})        
    \end{bmatrix}
    $}
\end{equation}

$\text{CORR}_{mp} \in \mathbb{R}^{m \times n}$ is the calculated CORAL-Correlation matrix. It captures the essence of aligning the feature distributions of unlabeled and labeled data in the projected feature space. In a similar approach, one can derive the similarity distribution $CORR_{ap}$ between \(z_a\) and \(z_p\) by substituting $z_m$ with $z_a$.

To ensure CORAL-Correlation Consistency across both branches, we apply cross-entropy loss $\ell_{ce}$ as the regularization mean, thereby aligning the CORAL-Correlation Consistency outputs. The formulation of loss $l_{d}$ can be conducted as follows:
\begin{equation}
    l_d = \frac{1}{m} \sum \ell_{ce} (CORR_{mp}, CORR_{ap})
\end{equation}

\subsection{Dynamic Feature Pool}
In extracting various features to confront the ambiguous segmentation boundaries, such as inconsistently dense areas and outcurved areas, we devise a Dynamic Feature Pool (DFP) $P$, for anchor point embeddings and iterative CCM computation. $P$ initializes $N$ slots per labeled training instance, dynamically updating anchor point embeddings with labeled features projected by $g_m$ and $g_a$. DFP selects embeddings only from positions where both $E_{main}$ and $E_{aux}$ yield correct predictions, ensuring feature reliability. It uses fusion method to adapt smoothly to variations in sample distribution by averaging old and new features with adjustable weights. Following \cite{He_2020_CVPR}, $P$ refreshes its slots for the labeled samples in the current mini-batch using a query-like mechanism, enabling dynamic and smooth updates of anchor point embeddings. This approach effectively aligns feature distributions between labeled and unlabeled data, enhancing the network's segmentation performance in accurately segmenting ambiguous boundaries.

\subsubsection*{Anchor Point Sampling Strategy}
Dynamic Feature Pool reaches a balance between computational efficiency and model performance by avoiding calculations across all labeled and unlabeled data. A confidence-based strategy is introduced for the selective extraction of features from unlabeled data. Initially, pixels with matching predictions from $E_{main}$ and $E_{aux}$ are identified, and $z_m$ and $z_a$ are sampled. For each class, pixels are ranked by confidence scores, and features from the top $i$ pixels are selected to represent the unlabeled data, with the total number of selected features $s = i \times C$, where $C$ is the number of classes. To enhance the diversity of anchor point embeddings, a strategy of randomly choosing $j$ embeddings per class from the feature pool $P$ is adopted. Moreover, simply selecting features with high confidence may lead to over-fitting, preventing the model from learning the true and comprehensive characteristics of the data. Therefore, a portion of features with low confidence is selected randomly. This sampling strategy results in $n = j \times C$ embeddings selected in total, which not only enriches the diversity but also enhances the generalization ability of the model, which helps the model to better select anchor points for alignment.

\section{Experiments and Results}
\subsection{Experiments}
\subsubsection{Dataset}
We evaluate our approach using the Left Atrium (LA) dataset \cite{xiong2021global} consisting of 100 gadolinium-enhanced MRI scans with corresponding ground truth masks. All scans have an isotropic resolution of $0.625^{3} \text{mm}$. Following \cite{yu2019uncertainty}, we split the dataset into 80 scans for training and 20 for testing.

\subsubsection{Implementation Details} 

CORN is implemented in PyTorch 2.4.0 with CUDA 12.1 on a single NVIDIA A100 GPU. During training, random cropping extracts $112 \times 112 \times 80$ sub-volumes according to \cite{yu2019uncertainty}, and V-Net \cite{7785132} is used as the backbone. Models are trained in batches of 4 (equal split of labeled and unlabeled data) using SGD with a learning rate of $0.01$, momentum of $0.9$, and weight decay of $1 \times 10^{-4}$ for  $1.5 \times 10^4$ iterations. A time-dependent Gaussian function $\lambda(t) = \beta \times e^{(-5 \times 1 - t/t_{\text{max}})^2}$ where $\beta = 1.00$, adjusts $\lambda_{c}$ and $\lambda_{d}$, both set to 0.1 initially. CCM module parameters are set to $c' = 64$, $j = 256$, and $i = 12800$. Inference uses a sliding window strategy without post-processing.

\subsubsection{Quantitative Evaluation and Comparison}
CORN is evaluated by four metrics: Dice(\%) $\uparrow$, Jaccard (\%) $\uparrow$, $95\%$ Hausdorff Distance (95HD) $\downarrow$ and Average Surface Distance (ASD)$\downarrow$. The experimental results are displayed in Table \ref{LA_results}. 

From Table I it can be seen that: (1) CORN achieves the best performance on all metrics at different ratio settings. For example, in the experiment with 5\% / 10\% / 20\% labeled data, CORN produces impressive performance gains of all metrics. Especially when the total amount of labeled data accounts for 20\%, CORN achieves 91.22\% Dice scores, which is very close to the fully supervised model. (2) Significant improvement occurs in 95HD and ASD(Average Surface Distance) metrics, i.e., with only 5\% labeled data, 95HD metric is 1.31\% lower than previous state-of-the-art model and ASD metric is 1.00\%.

These experiment results indicate that CORN can effectively align unlabeled data with labeled data, and improve the feature representation, which avoids inaccurate supervision. Thus, CORN obtains a stunning improvement. Meanwhile, the visualized results shown in Fig. \ref{Fig.visual} present that CORN can be effectively aware of both most global shape and local organ details, which remarkably enhances the performance on ambiguous boundaries and prevents the occurrence of ``Segmentation Breakage''. This ability of CORN has also been demonstrated in the enhancement of 95HD metric and ASD metric. 95HD measures the distance between the model's segmentation results and the ground truth (GT) in the worst-case scenario, a lower value indicates a reduced impact of extreme outliers. Similar to 95HD metric, ASD (Average Surface Distance) evaluates the average segmentation accuracy, indicating a more similar shape and details to GT.

\begin{table}[htbp]
\centering
\caption{Comparison with state-of-the-art methods on the LA dataset. }
\label{LA_results}
\resizebox{\columnwidth}{!}{%
\begin{tabular}{c|cc|cccc}
\hline 
\toprule
\multirow{2}{*}[-0.4mm]{\textbf{Method}} & \multicolumn{2}{c|}{\textbf{Scans used}} & \multicolumn{4}{c}{\textbf{Metrics}} \\ 
 & Labeled & Unlabeled & Dice(\%) $\uparrow$ & Jaccard(\%) $\uparrow$ & 95HD (voxel) $\downarrow$ & ASD (voxel) $\downarrow$ \\ 
\hline
\midrule
V-Net & 4 & 0 & 43.32& 31.43 & 40.91 & 12.13 \\ 
V-Net & 8 & 0 & 79.87 & 67.60 & 26.65 & 7.94\\ 
V-Net & 16 & 0 & 85.94 & 75.99 & 16.70 & 4.80\\ 
V-Net & 80 & 0 & 90.98 & 83.61 & 8.58 & 2.10 \\ 
\hline
DTC \cite{Luo_Chen_Song_Wang_2021} (AAAI'21) & \multirow{5}{*}{4 (5\%)} & \multirow{5}{*}{76 (95\%)} & 80.14 & 67.88 & 24.08 & 7.18 \\ 
SS-Net \cite{wu2022exploring} (MICCAI'22) &  &  & 83.33 & 71.79 & 15.70 & 4.33 \\ 
MC-Net+ \cite{wu2022mutual} (MedIA'22) &  &  & 83.23 & 71.70 & 14.92 & 3.43 \\ 
CAML \cite{gao2023correlationaware} (MICCAI'23) &  &  & \underline{86.78} & \underline{75.97} & \underline{10.34} & \underline{2.89} \\  
\textbf{Ours} &  &  & \textbf{88.23} & \textbf{79.12} & \textbf{9.51} & \textbf{1.73} \\ 
\hline

DTC \cite{Luo_Chen_Song_Wang_2021} (AAAI'21) & \multirow{6}{*}{8 (10\%)} & \multirow{6}{*}{72 (90\%)} & 84.55 & 73.91 & 13.80 & 3.69 \\ 
SS-Net \cite{wu2022exploring}  (MICCAI'22) &  &  & 86.56 & 76.61 & 12.76 & 3.02 \\ 
MC-Net+ \cite{wu2022mutual} (MedIA'22) &  &  & 87.68 & 78.27 & 10.35 & \underline{1.85} \\ 
CE-MT \cite{CE-MT} (BIBM'23) & & & 87.34 & 77.91 & 8.79 & 2.12 \\
CAML \cite{gao2023correlationaware} (MICCAI'23) &  &  & \underline{89.33} & \underline{80.17} & \underline{8.82} & 2.08 \\ 
\textbf{Ours} &  &  & \textbf{89.70} & \textbf{81.45} & \textbf{7.26} & \textbf{1.62} \\ 
\hline

DTC \cite{Luo_Chen_Song_Wang_2021} (AAAI'21) & \multirow{6}{*}{16 (20\%)} & \multirow{6}{*}{64 (80\%)} & 87.79 & 78.52 & 10.29 & 2.50 \\ 
SS-Net \cite{wu2022exploring}  (MICCAI'22) &  &  & 88.19 & 79.21 & 8.12 & 2.20 \\ 
MC-Net+ \cite{wu2022mutual} (MedIA'22) &  &  & 90.60 & 82.93 & 6.27 & \underline{1.58} \\ 
CE-MT \cite{CE-MT} (BIBM'23) & & & 89.67 & 80.53 & 7.21 & 1.99 \\
CAML \cite{gao2023correlationaware}  (MICCAI'23) &  &  & \underline{90.37} & \underline{82.56} & \underline{5.64} & 1.72 \\ 
\textbf{Ours} &  &  & \textbf{91.22} & \textbf{83.96} & \textbf{5.34} & \textbf{1.54} \\ 
\hline 
\bottomrule
\end{tabular}%
}
\end{table}

\subsection{CORAL Loss and Domain Alignment}
The CORAL (Correlation Alignment) loss is designed to align the statistical distributions of features between source (labeled data) and target (unlabeled data) domains. This loss function minimizes the difference in covariance matrices between the two domains, promoting feature distributions. The theoretical foundation is based on the idea that minimizing the distance between the covariance matrices of source and target domains helps domain shift, thereby improving the model's ability to generalize. We evaluate the alignment by examining the loss function. The convergence of the CORAL loss function is indicative of effective domain alignment. As the loss decreases, the statistical distributions of features in the source and target domains become more similar, suggesting successful alignment.
\begin{figure}[htbp] 
\includegraphics[width=0.99\columnwidth]{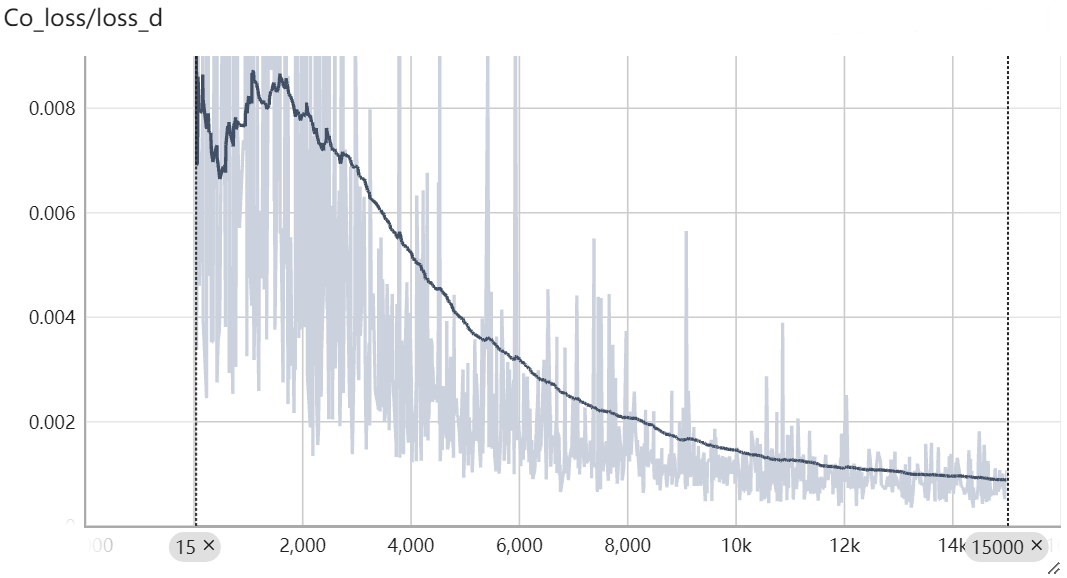} 

\caption{The image of CORAL loss curves over training epochs.}
\label{Fig.coral_loss} 
\end{figure}
Fig \ref{Fig.coral_loss} shows the loss curves over training epochs, where a steady decrease in CORAL loss correlates with improved domain alignment. This alignment is critical for enhancing model performance in domain adaptation scenarios, as evidenced by the experimental results.

\subsection{Ablation Study}
In the ablation study, our focus is to evaluate the effectiveness of the CORN network, with MC-Net \cite{Xie_2020_CVPR}  serving as the baseline.
Given that the DFP module and the CCM module are connected in series and DFP is a pivotal component for the CCM to filter and recommend the high-quality anchor points, the study to check the effectiveness of DFP is structured to compare the CORAL-correlation Consistency computation method to the traditional cosine similarity method. And the method to check the CCM module is to decrease the weight of CORAL loss and replace the proposed filter and fusion strategy with random sampling and no fusion process that indirectly eliminates the effects of DFP. Table \ref{ablation} demonstrates the experiment outcomes. It can be seen that, by combining both DFP and CCM modules, CORN achieves a massive advancement of 13.25\% in Dice metric, 13.42\% in 95HD metric, and 5.28\% in ASD metric with only 5\% labeled data. Furthermore, the ablation study has demonstrated that every single module is effective, \textit{i.e.}, Baseline + CCM shows a 12.7\% enhancement in 95HD with 5\% labeled data, demonstrating the ability of CCM to capture global and local features. The ablation study shows that the CORAL method using second-order statistics information (covariance) of labeled and unlabeled feature distributions has more advantages in extracting global and local features compared to cosine similarity, which only focuses on pixel-level information.

\begin{table}[htbp]
\centering
\caption{Ablation study of our proposed CORN on the LA dataset.}
\label{ablation}
\resizebox{\columnwidth}{!}{
\begin{tabular}{cc|ccc|cccc}
\hline 
\toprule
\multicolumn{2}{c|}{\textbf{Scans used}} & \multicolumn{3}{c|}{\textbf{Components}} & \multicolumn{4}{c}{\textbf{Metrics}} \\
Labeled & Unlabeled & Baseline & CCM & DFP& Dice (\%) $\uparrow$ & Jaccard (\%) $\uparrow$ & 95HD (voxel) $\downarrow$ & ASD (voxel) $\downarrow$ \\ \hline
\midrule
\multirow{4}{*}{4 (5\%)} & \multirow{4}{*}{76 (95\%)} & $\checkmark$ & & & 74.79 & 61.91 & 22.93 & 7.01 \\
& & $\checkmark$ & $\checkmark$ & & \underline{87.67} & 78.24 & \underline{10.23} & \underline{1.83} \\ 
& & $\checkmark$ & & $\checkmark$ & 87.74 & \underline{78.33} & 10.82 & 1.95 \\
& & $\checkmark$ & $\checkmark$ & $\checkmark$ & \textbf{88.23} & \textbf{79.12} & \textbf{9.51} & \textbf{1.73} \\
\hline

\multirow{4}{*}{8 (10\%)} & \multirow{4}{*}{72 (90\%)} & $\checkmark$ & & & 79.75 & 68.82 & 15.74 & 4.27 \\
& & $\checkmark$ & $\checkmark$ & & \textbf{89.79} & \textbf{81.62} & 7.41 & 1.73 \\ 
& & $\checkmark$ & & $\checkmark$ & 89.64 & 81.42 & 7.57 & 1.78 \\
& & $\checkmark$ & $\checkmark$ & $\checkmark$ & \underline{89.70} & \underline{81.45} & \textbf{7.26} & \textbf{1.62} \\ 
\hline
\multirow{4}{*}{16 (20\%)} & \multirow{4}{*}{64 (80\%)} & $\checkmark$ & & & 86.50 & 77.50 & 10.90 & 2.91 \\
& & $\checkmark$ & $\checkmark$ & & 90.91 & 83.47 & 5.73 & \textbf{1.51} \\ 
& & $\checkmark$ & & $\checkmark$ & 89.98 & 82.56 & 5.94 & 1.59 \\
& & $\checkmark$ & $\checkmark$ & $\checkmark$ & \textbf{91.22} & \textbf{83.96} & \textbf{5.34} & \underline{1.54} \\
\bottomrule
\hline
\end{tabular}
}
\end{table}

\section{Conclusion}
\sloppy
In this paper, we proposed the CORAL-Correlation Consistency Network (CORN) focusing on Left Atrium semi-supervised segmentation, enabling model be aware of both global shape and local organ details. The core innovations of CORN involve the Dynamic Feature Pool (DFP) to guarantee proper feature representation to avoid inaccurate supervision and utilization and the CORAL-Correlation Consistency Module (CCM) to leverage second-order statistical information for feature consistency. Extensive experimental analysis and visualizations on the LA dataset \cite{xiong2021global} demonstrate a notable improvement beyond current state-of-the-art (SOTA) results, demonstrating its ability to avoid ``Segmentation Breakage'' and provide better boundary details.

\section*{Acknowledgement}
Our work was supported in part by the Guangdong Provincial Key Laboratory IRADS (2022B1212010006, R0400001-22)  and in part by Guangdong Higher Education Upgrading Plan (2021-2025) with UIC research grant UICR0400025-21

\bibliographystyle{IEEEtran}
\bibliography{BIBMbib}

\end{document}